\pdfoutput=1

\documentclass[11pt]{article}

\usepackage{acl}

\usepackage{times}
\usepackage{latexsym}

\usepackage[T1]{fontenc}
\usepackage{pgfplots} 

\usepackage[utf8]{inputenc}

\usepackage{multirow}
\usepackage{amsmath}
\usepackage{capt-of}
\usepackage{tabularx}
\usepackage{epsfig}
\usepackage{amssymb}
\usepackage{amsfonts}
\usepackage{booktabs}
\usepackage{scalerel}
\usepackage[inline]{enumitem}
\usepackage{listings}
\usepackage{varwidth}
\usepackage[export]{adjustbox}
\usepackage{tikz}
\usetikzlibrary{tikzmark}
\usepackage{cleveref}

\usepackage{stmaryrd}
\usepackage{bbm}

\usepackage{algorithm}
\usepackage[noend]{algpseudocode}

\definecolor{deepblue}{rgb}{0,0,0.5}
\definecolor{officeblue}{RGB}{0,102,204}
\definecolor{deepred}{rgb}{0.6,0,0}
\definecolor{deepgreen}{rgb}{0,0.5,0}
\definecolor{mybrickred}{RGB}{182,50,28}

\definecolor{fillcolor}{RGB}{216,217,252}

\algnewcommand\algorithmicrequireb{{\hspace{0.85cm}}}
\algnewcommand\INPTDESCB{\item[\algorithmicrequireb]}

\algnewcommand\algorithmicfuncdesc{\textbf{Function:}}
\algnewcommand\FUNCDESC{\item[\algorithmicfuncdesc]}
\algnewcommand\algorithmicfuncdescb{{\hspace{1.48cm}}}
\algnewcommand\FUNCDESCB{\item[\algorithmicfuncdescb]}
\algnewcommand{\algorithmicgoto}{\textbf{goto}}
\algnewcommand{\Goto}[1]{\algorithmicgoto~\ref{#1}}

\usepackage{amsmath,amsfonts,bm}

\def\eqref#1{equation~\ref{#1}}

\def\1{\bm{1}}
\newcommand{\train}{\mathcal{D}}

\def\vtheta{{\bm{\theta}}}

\def\vg{{\bm{g}}}
\def\vh{{\bm{h}}}

\def\vp{{\bm{p}}}
\def\vq{{\bm{q}}}

\def\vu{{\bm{u}}}

\def\vx{{\bm{x}}}

\DeclareMathAlphabet{\mathsfit}{\encodingdefault}{\sfdefault}{m}{sl}
\SetMathAlphabet{\mathsfit}{bold}{\encodingdefault}{\sfdefault}{bx}{n}

\newcommand{\Ls}{\mathcal{L}}

\newcommand\our{\textsc{Xpr}}
\newcommand\clwe{\textsc{Clwe}}
\newcommand\plm{\textsc{Clse}}
\newcommand\dsname{WikiXPR}
\newcommand\xpco{\textsc{XpCo}}

\title{Cross-Lingual Phrase Retrieval}

\newcommand*\samethanks[1][\value{footnote}]{\footnotemark[#1]}

\author{Heqi Zheng$^{12}$\thanks{\ \ Co-first authors with equal contributions.},~~Xiao Zhang$^{1}\samethanks[1]$,~~Zewen Chi$^{1}\samethanks[1]$,~~Heyan Huang$^{12}$\\
\textbf{Tan Yan}$^{1}$\textbf{,}~~\textbf{Tian Lan}$^{1}$\textbf{,}~~\textbf{Wei Wei}$^{3}$\textbf{,}~~\textbf{Xian-Ling Mao}$^{1}$\thanks{\ \ Corresponding author.}\\
$^{1}$School of Computer Science and Technology, Beijing Institute of Technology\\
$^2$Beijing Engineering Research Center of High Volume Language Information Processing\\
and Cloud Computing Applications\\
$^3$Huazhong University of Science and Technology\\
\texttt{\{hikiz,yotta,czw,hhy63,yt,lt,maoxl\}@bit.edu.cn,weiw@hust.edu.cn}\\}

\usepackage{pgfplots}
\pgfplotsset{compat=1.17}

\begin{document}
\maketitle
\begin{abstract}

Cross-lingual retrieval aims to retrieve relevant text across languages. Current methods typically achieve cross-lingual retrieval by learning language-agnostic text representations in word or sentence level. However, how to learn phrase representations for cross-lingual phrase retrieval is still an open problem. In this paper, we propose \our{}, a cross-lingual phrase retriever that extracts phrase representations from unlabeled example sentences. Moreover, we create a large-scale cross-lingual phrase retrieval dataset, which contains 65K bilingual phrase pairs and 4.2M example sentences in 8 English-centric language pairs. Experimental results show that \our{} outperforms state-of-the-art baselines which utilize word-level or sentence-level representations. \our{} also shows impressive zero-shot transferability that enables the model to perform retrieval in an unseen language pair during training. Our dataset, code, and trained models are publicly available at \url{github.com/cwszz/XPR/}.

\end{abstract}

\section{Introduction}



Phrase retrieval aims to retrieve relevant phrases from a large phrase set, which is a critical part of information retrieval. Recent studies on phrase retrieval learn dense representations of phrases, and achieve promising results in entity linking, slot filling, and open-domain question answering tasks \cite{gillick-etal-2019-learning,learning-dense,PhraseRL}.
Nonetheless, most of the studies focus on monolingual scenarios, leaving the cross-lingual phrase retrieval unexplored.

Various methods have been proposed to perform cross-lingual text retrieval, which learns cross-lingual word or sentence representations shared across languages. Cross-lingual word representation methods typically train word embeddings on each language separately, and then learn an embedding mapping between the embedding spaces of different languages \cite{mikolov2013exploiting,dinu2014improving}. Then, the bilingual word pairs can be retrieved between vocabularies using nearest neighbor search, which is also known as bilingual lexicon induction \cite{vecmap,muse}.
Cross-lingual sentence retrieval is typically achieved by learning a sentence encoder on multilingual text corpora with self-supervised pretraining tasks \cite{xlm,xlmr}, or large-scale parallel corpora \cite{artetxe-schwenk-2019-massively-laser}, or both \cite{infoxlm}. The trained sentence encoders produce language-agnostic sentence representations, which enables sentences to be retrieved across languages.

Despite the effectiveness of word-level and sentence-level methods, how to learn phrase representations for cross-lingual phrase retrieval is still an open problem. Learning cross-lingual phrase representations is challenging in two aspects.
First, a phrase is a conceptual unit containing multiple words, so it is necessary to model the interaction between words, which is not considered in word-level methods.
Second, a phrase contains fewer words with less information compared to sentences, which prevents sentence encoders from taking the advantage of the ability of understanding full-length sentences.

Thus, in this paper, we propose a novel cross-lingual phrase retriever named as \our{}. Unlike previous cross-lingual retrieval methods that directly encode the input text, \our{} produces phrase representations using example sentences, which can be collected from unlabeled text corpora.
Initialized with a pretrained cross-lingual language model, \our{} can either directly serve as an unsupervised retriever, or be further trained to produce better-aligned phrase representations.
Besides, we propose the cross-lingual phrase contrast (\xpco{}) loss for training \our{}, where the model is trained to distinguish bilingual phrase pairs from negative examples. Furthermore, we create a cross-lingual phrase retrieval dataset, namely \dsname{}. \dsname{} contains 65K bilingual phrase pairs of eight language pairs, and provides example sentences for each phrase. 

We conduct a comprehensive evaluation of \our{} on \dsname{} under four evaluation settings, i.e., unsupervised, supervised, zero-shot transfer, and multilingual supervised. Our \our{} model substantially outperforms the retrieval baselines based on cross-lingual word embeddings and cross-lingual sentence encoders. \our{} also shows impressive zero-shot transferability that enables the model to be trained in a language pair and directly perform phrase retrieval for other language pairs. Moreover, we present an in-depth analysis on \our{}, showing that using example sentences improves both the learned \our{} model and the phrase representations.

Our contributions are summarized as follows:
\begin{itemize}
\item We propose \our{}, a novel cross-lingual phrase retriever that utilizes example sentences to produce phrase representations.
\item We propose the cross-lingual phrase contrast loss for training \our{}.
\item We demonstrate the effectiveness of \our{} on eight language pairs under four evaluation settings.
\item We create a cross-lingual phrase retrieval dataset, which provides 65K bilingual phrase pairs with 4.2M example sentences in 8 language pairs. 
\end{itemize}

\section{Related Work}

\paragraph{Cross-Lingual Retrieval}

Current cross-lingual text retrieval methods focus on word-level and sentence-level scenarios. Word-level cross-lingual retrieval methods typically train word embeddings on each language separately, and then align the word embeddings between languages by learning a mapping function \cite{mikolov2013exploiting,dinu2014improving,vecmap-supervised-artetxe2016emnlp,vecmap,muse,improve_cl_word_embedding,muse-retreival-criterion-joulin-etal-2018-loss}.
Similarly, cross-lingual sentence retrieval can be achieved by aligning sentence representations across different languages. LASER~\cite{artetxe-schwenk-2019-massively-laser} learns a multilingual auto-encoder on multilingual parallel corpora to produce language-agnostic sentence embeddings. Training on parallel corpora, cross-lingual sentence representations can also be learned with neural machine translation~\cite{laser-schwenk-2018-filtering}, contrastive learning~\cite{m-use,LABSE,infoxlm}, translation span corruption~\cite{mt6}, or knowledge distillation~\cite{ham-kim-2021-semantic-alignment}. Thanks to the recent language model pretraining technique~\cite{bert}, sentence encoders can be learned on a multilingual unlabeled text corpus without using parallel corpora \cite{xlm,xlmr,xlme,goswami-etal-2021-cross}.

\paragraph{Phrase Retrieval}
Recent research on phrase retrieval typically learns phrase representations. \citet{seo-etal-2019-real-phrase-dense-qa} propose to treat phrases as the smallest retrieval unit for open-domain question answering, where the phrases are encoded as indexable query-agnostic representations. The retrieval methods can be further improved with self-supervised pretraining, leading to better performance on open-domain question answering \cite{learning-dense,PhraseRL}. Additionally, DEER~\cite{gillick-etal-2019-learning} formulates the entity linking task as an entity phrase retrieval problem. However, these works study phrase retrieval in a monolingual scenario while we focus on cross-lingual phrase retrieval.

\paragraph{Contrastive Learning}
Contrastive learning learns representations by a contrastive loss that encourages the positive data pairs to be more similar than other data pairs.
It has shown to be effective for learning representations of a wide range of modalities including visual representations~\cite{Moco,simclr,byol}, sentence representations~\cite{Kong2020A,infoxlm,gao-etal-2021-simcse}, audio representations~\cite{saeed2021contrastive}, etc.
Different from previous work that performs contrastive learning at sentence level, we introduce contrastive learning to learn phrase representations. 


\begin{figure*}[t]
\centering
\includegraphics[width=0.9\textwidth]{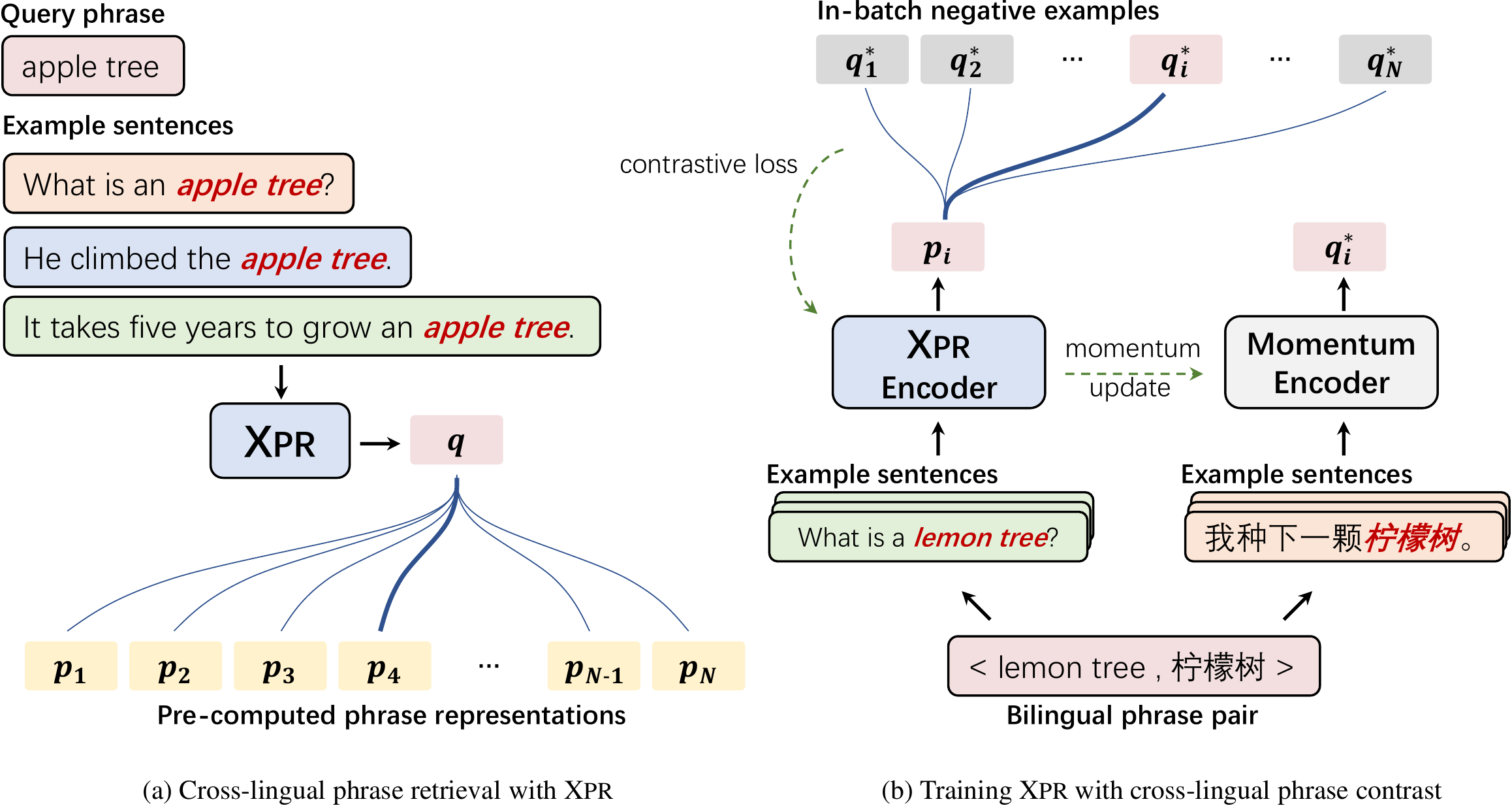}
\caption{The overview of \our{}. (a) The cross-lingual phrase retrieval procedure of \our{}. \our{} first extracts phrase representations from the example sentences sampled from an unlabeled text corpus, and then performs cross-lingual phrase retrieval with nearest search. (b) The cross-lingual phrase contrast (\xpco{}) loss. Notice that the example sentences are sampled separately for each phrase, i.e., training without parallel sentences.}
\label{fig:xpr}
\end{figure*}

\section{Methods}
\label{sec:method}

Figure~\ref{fig:xpr} shows the overview of \our{}. In this section, we first introduce the model architecture of \our{}, and then present the cross-lingual phrase contrast loss. Finally, we show the training procedure of \our{}.


\subsection{Model Architecture}
\label{sec:modelarch}

The model architecture of \our{} is a Transformer~\cite{transformer} encoder shared across different languages. \our{} can be initialized with pretrained cross-lingual language models, which have shown to produce well-aligned sentence representations \cite{xtreme,infoxlm}.

Given a phrase $p$ and an example sentence $x = w_1, \dots, w_n$ with $n$ tokens that contain the phrase. We denote the start and end indices of $p$ as $s$ and $e$, i.e., $p = w_s, \dots, w_e$. \our{} first encodes $x$ into a sequence of contextualized token representations\footnote{Following \citet{infoxlm}, we take the hidden vectors from a specific hidden layer as the token representations rather than only the last layer.}
\begin{align}
    \vh_1, \dots, \vh_n = \text{Transformer}(w_1, \dots, w_n).
\end{align}
Then, the phrase is represented as the average of the phrase tokens 
\begin{align}
    \vx = \frac{1}{e - s + 1} \sum_{i=s}^{e} \vh_i.
\end{align}
In general, a phrase can have more than one example sentence. Considering $m$ example sentences $\mathcal{X} = x_1, \dots, x_m$ for the phrase $p$, \our{} encodes the sentences separately, and uses the average of the phrase representations as the final phrase representation, i.e., $\sum_{x \in \mathcal{X}} \vx / m$.
Notice that \our{} does not introduce additional parameters beyond the original Transformer encoder. Thus, after the initialization with a pretrained cross-lingual language model, \our{} can directly serve as an unsupervised cross-lingual phrase retriever.

\subsection{Cross-Lingual Phrase Contrast Loss}


Recent work \cite{simclr,Kong2020A} has demonstrated the effectiveness of contrastive learning framework for learning visual and text representations. 
To learn language-agnostic phrase representations, we propose the cross-lingual phrase contrast (\xpco{}) loss, where the goal is to distinguish the bilingual phrase pairs from negative examples.

Formally, consider a mini-batch $\mathcal{B} = \{ \mathcal{P}, \mathcal{Q}\}$ of bilingual phrase pairs, where $\mathcal{P} = \{ p \}^N$ and $\mathcal{Q} = \{ q \}^N$ stand for $N$ phrases in a language and their translations in another language, respectively. For each phrase $p \in \mathcal{P}$, we sample example sentences $\mathcal{X}$ for $p$, and compute the phrase representation $\vu$ as described in Section~\ref{sec:modelarch}.
Following \citet{simclr}, we apply a projection head over $\vu$ that consists of two linear layers with a ReLU in between and a $l_2$ normalization followed. For simplicity, we denote the above operation that converts an input phrase $p$ to a normalized vector as
$\vp = f(p, \mathcal{X}; \vtheta)$
, where $\vtheta$ stands for the parameters of the encoder and the projection head.
For each phrase $q \in \mathcal{Q}$, we employ a momentum encoder \cite{Moco} to encode $q$:
$\vq^* = f(q, \mathcal{Y}; \vtheta_\text{m})$
, where $\mathcal{Y}$ represents the example sentences of $q$, and $\vtheta_\text{m}$ represents the parameters of the momentum encoder.

For the $i$-th phrase $p_i \in \mathcal{P}$, $q_i \in \mathcal{Q}$ is its corresponding positive example and the other $N-1$ phrases are treated as negative examples. The contrastive loss in the direction of $\mathcal{P} \rightarrow \mathcal{Q}$ is defined as
\begin{align}
    \Ls(\mathcal{P} \rightarrow \mathcal{Q}) = -\sum_{i=1}^{N} \log \frac{\exp(\vp_i^\top\vq_i^*/T)}{\sum_{j=1}^{N} \exp(\vp_i^\top\vq_j^*/T)}
\end{align}
Similarly, we employ an additional contrastive loss in the direction of $\mathcal{Q} \rightarrow \mathcal{P}$. The \xpco{} loss combines both directions, which is defined as
\begin{align}
    \Ls_\xpco{} = \Ls(\mathcal{P} \rightarrow \mathcal{Q}) + \Ls(\mathcal{Q} \rightarrow \mathcal{P})
\end{align}
where $T$ is the softmax temperature.

\begin{algorithm}[t]
\caption{Training procedure of \our{}}
\label{alg:xpr}
\begin{algorithmic}[1]
\Require Bilingual phrase pair corpus $\train$, unlabeled text corpus $\mathcal{U}$, learning rate $\tau$, momentum coefficient $\mu$
\Ensure \our{} parameters $\vtheta$
\State Initialize $\vtheta, \vtheta_\text{m}$
\While{not converged}
\State $(\mathcal{P},\mathcal{Q}) \sim \train$
\For {$i = 1,2,\dots,N$}
\State $\mathcal{X} \sim \mathcal{U}$ ~~s.t. $p_i \subset x$, $\forall x \in \mathcal{X}$
\State $\mathcal{Y} \sim \mathcal{U}$ ~~s.t. $q_i \subset y$, $\forall y \in \mathcal{Y}$
\State $\vp_i = f(p_i, \mathcal{X}; \vtheta)$
\State $\vp_i^* = f(p_i, \mathcal{X}; \vtheta_\text{m})$
\State $\vq_i = f(q_i, \mathcal{Y}; \vtheta)$
\State $\vq_i^* = f(q_i, \mathcal{Y}; \vtheta_\text{m})$
\EndFor
\State \textbf{end for}
\State $\vg \gets \nabla_\vtheta \Ls_\xpco{}$
\State $\vtheta \gets \vtheta - \tau \vg$
\State $\vtheta_\text{m} \gets \mu \vtheta_\text{m} + (1-\mu) \vtheta$
\EndWhile
\State \textbf{end while}
\end{algorithmic}
\end{algorithm}

\subsection{Training Procedure of \our{}}

Algorithm~\ref{alg:xpr} illustrates the training procedure of \our{}. We initialize the \our{} encoder $\vtheta$ and the momentum encoder $\vtheta_\text{m}$ with a pretrained cross-lingual language model. For each training step, we first sample a mini-batch of bilingual phrase pairs $(\mathcal{P}, \mathcal{Q})$ from the bilingual phrase pair corpus $\train$, and then sample example sentences $\mathcal{X}$ and $\mathcal{Y}$ for $\mathcal{P}$ and $\mathcal{Q}$, respectively. Each example sentence $x \in \mathcal{X}$ should contain the phrase $p_i$, which is denoted as $p_i \subset x$.
With the phrase representations produced by the two encoders, we compute the \xpco{} loss, and update $\vtheta$ with gradient descent. Notice that we do not perform back-propagation in the momentum encoder, which is learned by a momentum update \cite{Moco} with a momentum coefficient of $\mu$.

\subsection{Phrase Retrieval with \our{}}
Given a phrase set $\mathcal{P} = \{ p \}^N$ with $N$ candidate phrases , the goal is to find $p \in \mathcal{P}$ with the same meaning of a query phrase $q$. With the trained \our{} encoder $\vtheta$, we first sample example sentences candidate phrases and then compute their representations $\{ \vp \}^N$ with $f(\cdot;\vtheta)$. 
Then, for a query phrase $q$, we can find the corresponding phrase by:
\begin{align}
    \hat{p} = \arg \max_{p_i} \{ \vp_i^\top\vq \}
\end{align}
In practice, the representations of candidate phrases can be pre-computed for reuse. Moreover, although the example sentence number is limited during training, we can use more example sentences to obtain better phrase representation for retrieval.


\section{\dsname{}: Cross-Lingual Phrase Retrieval Dataset}
\label{sec:ds}
To evaluate our model,
we create \dsname{}, a cross-lingual phrase retrieval dataset extracted from Wikipedia. \dsname{} consists of bilingual phrase pairs in eight English-centric language pairs, and contains large-scale example sentences for the phrases, which enable models to leverage contextual information to better understand phrases. In what follows, we describe how we construct the \dsname{} dataset.

\subsection{Phrase Pair Mining}
Manually translating phrases is expensive when building a large-scale bilingual phrase pair corpus. Therefore, we leverage the link information within Wikipedia for mining bilingual phrase pairs. Specifically, we first extract inter-language linked wiki entries from dbpedia\footnote{\url{downloads.dbpedia.org/2014/en/}}. We treat English as the pivot language, and choose a range of diverse languages to build our datasets, so that the models can be evaluated with different language families and scripts. We filter out time expressions, and the phrase pairs with low edit distance using ROUGE-L~\cite{Lin2004ROUGEAP} as the distance measure. The phrase pairs with bidirectional ROUGE-L values higher than $0.5$ are removed.

\subsection{Example Sentence Retrieval}

In addition to phrase pairs in diverse languages, \our{} also provides example sentences for each phrase, which aims to facilitate the research on phrase representation learning with example sentences. For each phrase, we retrieve example sentences from an unlabeled text corpus. In specific, we first extract raw sentences from Wikipedia dumps as our unlabeled text corpus. Then, we build sentence indices with the Elasticsearch \footnote{\url{www.elastic.co}} searching engine. For each phrase, we 
retain the searched sentences with at least $10$ more characters than the phrase as the results. Besides, we only retain $32$ example sentences for each phrase to keep a reasonable size for the resulting example sentence corpus.

\begin{table}[t]
\centering
\small
\renewcommand\tabcolsep{3.0pt}
\scalebox{0.8}{
\begin{tabular}{lrrrrrrrrr}
\toprule
& \textbf{ar-en} &  \textbf{de-en} & \textbf{es-en} & \textbf{fr-en} & \textbf{ja-en} &  \textbf{ko-en} & \textbf{ru-en}  & \textbf{zh-en} &  \textbf{Total} \\ \midrule
Train & 4222 & 1931 & 1333 & 1315 & 14745 & 2138 & 5229 & 8326 & 39239 \\ 
Dev & 1408 & 644 & 445 & 438 & 4915 & 713 & 1743 & 2775 & 13081 \\ 
Test & 1407 & 644 & 445 & 438 & 4915 & 713 & 1743 & 2775 & 13080 \\ 
\bottomrule
\end{tabular}
}
\caption{The number of bilingual phrase pairs for each language pair in \dsname{}.}
\label{table:dataset}
\end{table}

\subsection{The Resulting \dsname{} Dataset}
As shown in Table~\ref{table:dataset}, we present the number of bilingual phrase pairs for each language pair in \dsname{}.
The resulting \dsname{} dataset consists of 65,400 phrase pairs in eight language pairs, and 4.2M example sentences in total. For each phrase, \dsname{} provides 32 example sentences extracted from Wikipedia text.
\dsname{} is split into training, dev, and test sets by 3:1:1, so \dsname{} can be used for diverse evaluation settings including the supervised setting, cross-lingual zero-shot transfer, etc. See detailed statistics in Appendix~\ref{appendix:ds}.

\section{Experiments}

In this section, we first present four evaluation settings for cross-lingual phrase retrieval, and describe the models to be compared. Then, we present the experimental results.
\subsection{Evaluation Settings}

We conduct experiments on the cross-lingual phrase retrieval task on our \dsname{} dataset. Detailed description of \dsname{} can be found in Section~\ref{sec:ds}. 
Since collecting or annotating parallel sentences can be expensive especially for low-resource languages, we only consider unlabeled text and the bilingual pairs provided by \dsname{} in our experiments.
According to the difference in the training resource, we present the following four evaluation settings.

\paragraph{Unsupervised} Under the unsupervised setting, the retrieval model should not use any bilingual phrase pairs or other cross-lingual supervision such as bilingual dictionaries and parallel corpus. The language representations are typically learned from unlabeled text corpora.

\paragraph{Supervised} In the supervised setting, the retrieval model is trained on and tested on bilingual phrase pairs for each language pair separately, e.g., training and testing with English-French phrase pairs.

\paragraph{Zero-Shot Transfer} Zero-shot transfer is a widely-used setting in cross-lingual understanding tasks \citet{xlm,wu2019beto}, where models are trained in a source language but evaluated on other languages. We introduce this setting to the cross-lingual phrase retrieval task, e.g., training a model with English-French phrase pairs but performing retrieval between English and Chinese phrases.

\paragraph{Multilingual Supervised} In this setting, the retrieval model is able to use training data in multiple languages, e.g., training a model using a combined training set over all languages in \dsname{} and testing it for each language.

\begin{table*}
\centering
\scalebox{0.8}{
\begin{tabular}{lrrrrrrrrr}
\toprule
\textbf{Model} & \textbf{ar-en}& \textbf{de-en} & \textbf{en-es} &\textbf{en-fr} & \textbf{en-ja} & \textbf{en-ko}& \textbf{en-ru} & \textbf{en-zh}  & \textbf{Avg} \\ \midrule
\multicolumn{9}{l}{~~\textit{Unsupervised}} \\

\clwe{} & 2.74& 0.78 & 0.00 &1.02 & 0.34 & 0.28 & 1.32 & 0.12 & 0.83\\
\plm{} &  9.70 & 19.10 & 29.21 & 20.89 & 4.83 & 11.50 & 16.98 & 8.76  & 15.12\\
\our{}  & \bf 14.71 & \bf 28.96 & \bf 42.25 & \bf 39.38 & \bf 7.34 & \bf 15.22 &  \bf 24.24  & \bf 11.26 & \bf 22.92
\\ \midrule
\multicolumn{9}{l}{~~\textit{Supervised}} \\
\clwe{} &  56.14& 33.62 & 63.71 &51.26 & 31.62 & 50.14 & 38.67 & 30.02  & 44.40 \\
\plm{}  & 20.58 & 18.79 & 36.06 & 26.60 & 16.73 & 24.58 & 21.32 & 17.69  & 22.79 \\
\our{} & \bf 88.63& \bf 81.44 &\bf 84.53 & \bf 80.18 &\bf 87.32 &\bf 80.83 &\bf 91.00 & \bf 77.62  &\bf 83.94 \\ \midrule
\multicolumn{9}{l}{~~\textit{Zero-shot transfer}} \\
\clwe{} & 0.04 & 0.32 & 0.22 & 0.23 & 0.00 & 2.24 & 0.09 &  30.02  & 4.15 \\
\plm{} & 6.18 & 10.25 & 16.07 & 10.39 & 6.73 & 9.75 & 8.35 & 17.69 & 10.68 \\
\our{} & \bf 74.12& \bf 73.60 & \bf 82.54 & \bf 77.36 & \bf 73.04 & \bf 78.52 & \bf 79.10 & \bf 77.62  & \bf 76.99 \\ \midrule
\multicolumn{9}{l}{~~\textit{Multilingual supervised}} \\
\clwe{}{} & 12.33 & 1.87 & 6.63 & 3.77 & 18.46 & 4.00 & 9.84 &  11.19  & 8.51 \\
\plm{} & 11.98 & 19.64 & 29.44 & 21.58 & 11.91 & 14.73 & 18.01 &  14.50  & 17.72 \\

\our{} &  \textbf{91.90}& \textbf{82.76} & \textbf{90.79} &\textbf{85.16} & \textbf{90.16} & \textbf{88.22} & \textbf{93.09} & \textbf{86.47}  & \bf 88.57 \\
\bottomrule
\end{tabular}
}
\caption{Accuracy@1 on \dsname{} cross-lingual phrase retrieval under four evaluation settings. Results are averaged over three random seeds in both the xx$\rightarrow$en and en$\rightarrow$xx directions, where `xx' denotes one of the eight non-English languages.}
\label{table:main}
\end{table*}

\subsection{Baselines}
Considering the lack of methods for cross-lingual phrase retrieval, we develop the following two baselines in our experiments:



\paragraph{\textsc{Clwe}}
Cross-lingual word embeddings (\textsc{Clwe}) encode words from various languages into a shared embedding space.
For each word in a phrase, we first represent it with the pretrained fastText multilingual word vectors~\cite{grave2018learning}, and then map it to a shared embedding space via the \textsc{VecMap}\footnote{\url{github.com/artetxem/vecmap}}~\cite{vecmap} tool. Notice that \textsc{VecMap} can be applied to both unsupervised and supervised scenarios. Finally, the retrieval is achieved by the nearest search using an average word vector as the phrase representation.

\paragraph{\plm{}} Cross-lingual sentence encoders (\plm{}) produce language-agnostic sentence representations for the input text sequence. We use  XLM-R$_\text{base}$~\cite{xlmr} as the sentence encoder, which is pretrained on a large-scale multilingual text corpus. For the unsupervised setting, we use the averaged hidden vector from a specific middle layer as the phrase representation. For the other settings, we follow \citet{wang-etal-2019-cross}, which learns an orthogonal mapping between the feature spaces of the training phrase pairs. As LASER~\cite{artetxe-schwenk-2019-massively-laser} and LaBSE~\cite{LABSE} utilize parallel corpora, we do not use them in our experiments.

As for our model \our{} described in Section~\ref{sec:method}, we initialize \our{} with XLM-r$_\text{base}$~\cite{xlmr} for a fair comparison.
For each step, we use a batch of $256$ phrase pairs and $4$ example sentences for each phrase. The model is optimized with the Adam~\cite{adam} optimizer with a learning rate of $2 \times 10^{-5}$ for $100$ epochs. The learning rate is scheduled with $1\%$ warm-up steps and a linear decay during training.

\subsection{Experimental Results}
Table~\ref{table:main} compares the three cross-lingual phrase retrieval models on our \dsname{} dataset under four different evaluation settings. 

\paragraph{Unsupervised Results}
As present in Table~\ref{table:main}, \our{} obtains the best performance over all languages without any cross-lingual supervision, achieving an average accuracy@1 of 22.92. On the contrary, \clwe{} and \plm{} only obtain 0.83 and 15.12, respectively. It indicates that \our{} successfully leverage example sentences to produce better phrase representations.
Besides, the performance varies in different languages. We observe that the retrieval between English and European languages can be easier than other language pairs when using \plm{} and \our{}.
It is worth mentioning that \clwe{} and \plm{} are proven to be effective for bilingual lexicon induction and cross-lingual sentence retrieval, respectively \cite{muse,xtreme}. Nonetheless, they do not perform as well as on word or sentence level tasks, indicating that they are not directly applicable to cross-lingual phrase retrieval.

\paragraph{Supervised Results}
Under the supervised setting, \our{} achieves an average accuracy of 83.94, largely outperforming the other two models over all evaluation language pairs.
Comparing the results between the unsupervised and the supervised settings, all the three models greatly benefit from the training data. In particular, \our{} pushes the average result from 7.34 to 87.32 for the en-ja phrase retrieval. The results suggest that the bilingual phrase pairs can help to learn cross-lingual alignment for both word-level and sentence-level representations.
We find that using training data brings more gains for \clwe{} than \plm{}, showing that the contextualized phrase representations in \plm{} can be harder to align.

\paragraph{Zero-shot Transfer}
In zero-shot transfer, the models are trained using an en-xx dataset but evaluated on all language pairs. The table only presents the results of the model trained on en-zh data. Detailed results of other transfer directions can be found in Appendix~\ref{appendix:zs}.
Although the \our{} model only learns on en-zh training data, it performs surprisingly well on other languages. On en-es and en-ko, \our{} even produces comparable results to the results in the supervised setting.
Comparing the results to the unsupervised setting, \our{} pushes the average accuracy from 22.92 to 76.99. 
This demonstrates the strong cross-lingual transferability of \our{}, which allows our model to be applied to low-resource languages without training data.
On the contrary, \plm{} fails to leverage the en-zh training data for the retrieval in other languages, resulting in a consistent performance drop. 

\paragraph{Multilingual Supervised}
In the multilingual supervised setting, \our{} obtains the best results over all models and settings, achieving an average accuracy of 88.57. 
Compared to the supervised setting, using the combined training data leads to consistent improvement over all languages, which demonstrates that \our{} successfully leverage the supervision signals from both the same and different languages.

\begin{table}[t]
\centering
\small
\scalebox{0.95}{
\begin{tabular}{lcccc}
\toprule
\textbf{Model} & \textbf{en-fr} & \textbf{en-ko} & \textbf{en-ru} & \bf Avg \\ \midrule
\multicolumn{5}{l}{~~\textit{Supervised}} \\
\our{} & \bf 80.18 & \bf 80.83  & \bf 91.00 & \bf 84.00 \\
~~$-$Example Sentence & 60.84 & 63.39 & 83.39  & 69.21 \\
~~$-$Momentum Update& 78.54  & 80.43 & 90.50 & 83.16 \\
~~$-$Projection Head & 77.28 & 77.70 & 89.16 & 81.38\\ \midrule
\multicolumn{5}{l}{~~\textit{Zero-shot transfer}} \\
\our{} & \bf 77.36  & \bf 78.52 & \bf 79.10 & \bf 78.33 \\
~~$-$Example Sentence& 60.50 & 60.03 & 60.90 & 60.48\\
\bottomrule
\end{tabular}
}
\caption{Ablation results of \our{} on \dsname{} cross-lingual phrase retrieval. We report the average accuracy@1 scores that are averaged in both the xx$\rightarrow$en and en$\rightarrow$xx directions. In zero-shot transfer, the models are trained using the en-zh data but evaluated on three other language pairs.}
\label{table:ablation}
\end{table}

\begin{figure}[t]
\centering

\includegraphics[width=0.4\textwidth]{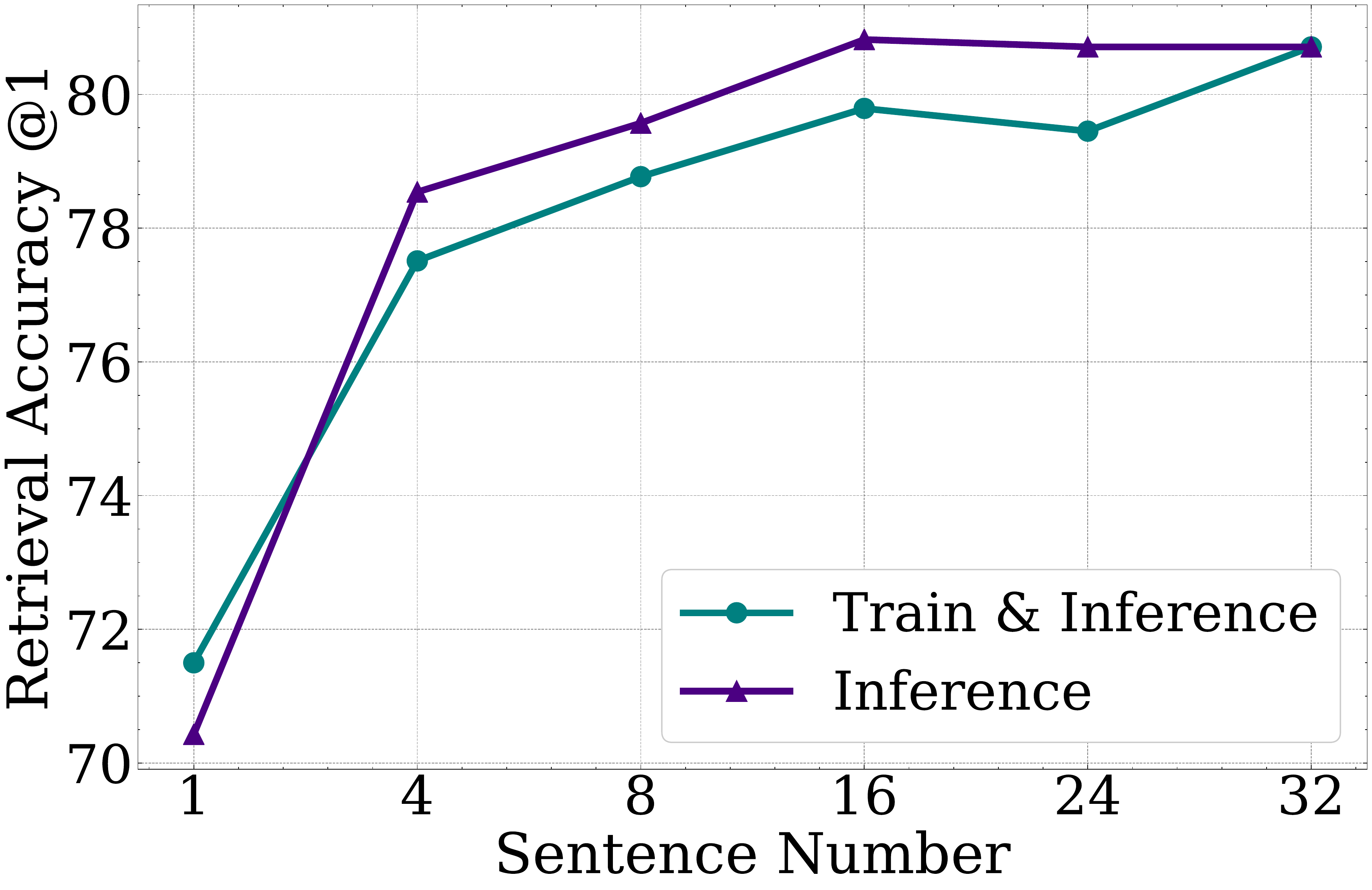}
\caption{Effects of the example sentence number. We train \our{} models on the en-fr set of \dsname{} under two settings: 1) Training and inference with various numbers of example sentences for each phrase, 2) Training with 32 example sentences for each phrase but inference with various numbers of example sentences.}
\label{fig:sentnum}
\end{figure}

\begin{table*}[t]
\centering
\scalebox{0.8}{
\begin{tabular}{lcccccccccccc}
\toprule
\textbf{Layer} & \textbf{1} & \textbf{2} & \textbf{3} &  \textbf{4} & \textbf{5} & \textbf{6} & \textbf{7} & \textbf{8} &  \textbf{9} & \textbf{10} & \textbf{11} & \textbf{12}\\ \midrule
\our{} (Unsupervised) & 29.22 & 29.22 & 31.39 &  35.73 & 36.64 & 37.67 & 35.50 & 36.07 & 35.84  & 37.10 & \bf 39.38 & 29.11 \\
\our{} (Supervised) & 30.94 & 30.06 & 36.45 & 42.13 & 48.13 & 51.52 & 51.29 & 55.82 & 59.51  & 70.85 & 78.01 & \bf 80.18 \\

\bottomrule
\end{tabular}
}
\caption{Effects of using phrase representations from different hidden layers of \our{}. We report the accuracy@1 on the en-fr set of \dsname{}, where the results are averaged in both the fr$\rightarrow$en and en$\rightarrow$fr directions.}
\label{table:layer}
\end{table*}

\begin{table}[t]
\centering
\scalebox{0.8}{
\begin{tabular}{lcccc}
\toprule
\textbf{Objective} & \textbf{en-fr} & \textbf{en-ko} & \textbf{en-ru} & \bf Avg \\ \midrule
\xpco{} & \bf 80.18 & \bf 80.83  & \bf 91.00 & \bf 84.00 \\
\textsc{MoCo} & 77.28 & 74.05 & 87.95  & 79.76 \\

\bottomrule
\end{tabular}
}
\caption{Comparison of \xpco{} and \textsc{MoCo} for training \our{}. We report the accuracy@1 on \dsname{}, where the results are averaged in both the xx$\rightarrow$en and en$\rightarrow$xx directions.}
\label{table:moco}
\end{table}

\subsection{Ablation Studies}
We conduct ablation studies by removing main components from \our{}. In specific, we compare three variants of \our{} that are trained without example sentences, momentum update, or projection head, respectively. The evaluation results are shown in Table~\ref{table:ablation}.

\paragraph{Example Sentence} We first investigate whether using example sentences helps cross-lingual phrase retrieval. During training, we remove the example sentences from \our{}, i.e., the model extracts the phrase representation only from the input phrase itself. As shown in Table~\ref{table:ablation}, removing example sentences substantially harms the performance of \our{} for both the supervised and zero-shot transfer settings. Notice that example sentences are not parallel across languages, but they still make the resulting phrase representations from different languages better aligned. Besides, compared to the supervised setting, the gains are even larger for zero-shot transfer, improving the average accuracy from 60.48 to 78.33. The above results demonstrate that using example sentences not only learns better phrase representations, but also encourages cross-lingual alignment.

\paragraph{Projection Head} We train a \our{} model without the projection head, i.e., directly using the average of the hidden vectors as the phrase representation. As shown in Table~\ref{table:ablation}, the projection head provides consistent gains on the three language pairs, showing the effectiveness of the projection head in contrastive learning. The results also agree with the finding in visual representation learning~\cite{simclr,chen2020big}.

\paragraph{Momentum Update} We study the effects of momentum update used in \our{}. It shows that the momentum update strategy slightly improves the results on all of the three evaluation language pairs, providing 0.84 accuracy improvement.

\subsection{Effects of Example Sentence Number}
We study the effects of the example sentence number used in \our{}. 
We conduct an evaluation on the en-fr set of \dsname{}, under two settings where the example sentence number varies during training or inference: 1) Training and inference with various numbers of example sentences for each phrase, 2) Training with 32 example sentences for each phrase but inference with various numbers of example sentences.

Figure~\ref{fig:sentnum} illustrates the evaluation results. 
It shows a trend that using more example sentences during inference notably improves the performance in both settings. The gain is larger when using fewer example sentences, demonstrating the effectiveness of using multiple example sentences for producing phrase representations. 
Comparing the results between the two settings, we find that the model moderately benefits from a large number of example sentences if we use a lower sentence number for inference. Although using more example sentences during training provides gains, the heavier computation load should be token into consideration.

\subsection{Effects of Layer}

Recent work \cite{infoxlm,xlme} has shown that a middle layer can produce better-aligned sentence representations than the last layer, resulting in higher cross-lingual sentence retrieval performance. We investigate which hidden layer of \our{} produces phrase representations that achieve higher retrieval accuracy. To this end, we evaluate \our{} using representations from various hidden layers on the en-fr set of \dsname{}.

As shown in Table~\ref{table:layer}, we present the evaluation results of \our{} under both the unsupervised and the supervised settings. For the unsupervised \our{}, we observe that Layer-11 produces the best results while the last layer even performs worse than the first layer. Differently, the supervised \our{} obtains the best results on Layer-12, indicating that our \xpco{} loss encourages the model to fully utilize the last few layers.
Moreover, it shows that using representations from higher layers of the supervised \our{} leads to consistent improvement.

\subsection{Comparison of Contrast Losses}
We explore whether using momentum contrast (\textsc{MoCo};~\citealt{Moco}) trains our \our{} model better, which is proven to be effective for cross-lingual language model pretraining \cite{infoxlm}. In specific, we train a variant of \our{} with \textsc{MoCo}, which maintains more negative examples encoded by the momentum encoder with a queue with a length of 1024.
The evaluation results are presented in Table~\ref{table:moco}. \xpco{} consistently outperforms \textsc{MoCo} on the three language pairs, suggesting that the negative examples stored in the queue can be out-of-date for contrastive learning.



\section{Conclusion}


In this work, we propose a cross-lingual phrase retriever \our{}, which outperforms the baseline retrievers on a range of diverse languages.
Moreover, we create a cross-lingual phrase retrieval dataset that contains diverse languages with large-scale example sentences. 
For future work, we would like to improve \our{} by: 1) extending \our{} to asymmetric retrieval scenarios such as open-domain question answering, 2) exploring how to utilize parallel corpora for training \our{}.

\section{Ethical considerations}

\our{} is designed as a cross-lingual phrase retriever that retrieve relevant phrases across different languages. We believe \our{} would help the communication between the people who speak different languages. Besides, our work can facilitate the research on multilingual natural language processing (NLP), which helps to build NLP applications for low-resource languages. In addition, we construct the \dsname{} dataset using open-source data from Wikipedia and dbpedia.

\section*{Acknowledgements}
The work is supported by National Key R\&D Plan (No. 2018YFB1005100), National Natural Science Foundation of China (No. U19B2020, 62172039, 61732005, 61602197 and L1924068),  the funds of Beijing Advanced Innovation Center for Language Resources (No. TYZ19005), and in part by CCF-AFSG Research Fund under Grant No.RF20210005, and in part by the fund of Joint Laboratory of HUST and Pingan Property \& Casualty Research (HPL). 
We would like to acknowledge Qian Liu for the helpful discussions.


\bibliographystyle{acl_natbib}
\bibliography{cc}

\newpage
\appendix

\section*{Appendix}

\section{Additional \dsname{} Statistics}
\label{appendix:ds}
Table~\ref{table:ds-detail} presents the detailed statistics of \dsname{}, including the number of phrase pairs, average phrase length, and the average length of example sentences.

\section{Detailed Results of Zero-Shot Transfer}
\label{appendix:zs}

Table~\ref{table:zsxpr} presents the evaluation results of \our{} on \dsname{} under the zero-shot transfer setting, where the \our{} model is trained in a source language pair but evaluated on target language pairs.

\begin{table*}[htbp]
\centering
\small
\renewcommand\tabcolsep{3.3pt}
\scalebox{0.77}{
\begin{tabular}{lcccccccc}
\toprule
&  \textbf{ar-en} & \textbf{de-en} & \textbf{en-es} &\textbf{en-fr} & \textbf{en-ja} & \textbf{en-ko} & \textbf{en-ru} & \textbf{en-zh} \\ \midrule
\multicolumn{9}{l}{~~\textit{Train}} \\
\#Phrase pairs &  4222 & 1931 & 1333 &1315 & 14745 & 2138 & 5229 & 8326  \\ 
Avg phrase length (xx/en) &  2.32 / 2.55 & 1.88 / 2.72 & 3.26 / 2.79 &3.05 / 2.81 & 7.73 / 2.48 & 5.18 / 2.35 & 1.95 / 2.55 & 4.47 / 2.48 \\
Avg example sentence length (xx/en) & 38.28 / 25.36 & 33.93 / 35.61 & 42.13 / 26.24 &34.75 / 27.63 & 116.57 / 32.01 & 130.63 / 25.18 & 32.26 / 27.56 & 97.30 / 33.25 
\\ \midrule
\multicolumn{9}{l}{~~\textit{Dev}} \\
\#Phrase pairs &  1408 & 644 & 445 &438 & 4915 & 713 & 1743 & 2775  \\ 
Avg phrase length (xx/en) &  2.32 / 2.55 & 1.89 / 2.64 & 3.31 / 2.72 &3.13 / 2.88 & 7.68 / 2.45 & 5.32 / 2.35 & 1.96 / 2.56 & 4.49 / 2.46 \\
Avg example sentence length (xx/en) &  38.04 / 24.58 & 34.10 / 35.53 & 43.02 / 26.74 &36.39 / 29.25 & 116.28 / 32.00 & 128.20 / 24.50 & 31.40 / 26.06 & 98.54 / 33.84
\\ \midrule
\multicolumn{9}{l}{~~\textit{Test}} \\
\#Phrase pairs & 1407 & 644 & 445 &438 & 4915 & 713 & 1743 & 2775 \\ 
Avg phrase length (xx/en) &  2.33 / 2.57 & 1.89 / 2.68 & 3.24 / 2.83 &3.04 / 2.75 & 7.75 / 2.47 & 5.13 / 2.34 & 1.93 / 2.56 & 4.48 / 2.50 \\
Avg example sentence length (xx/en) &  27.75 / 24.94 & 33.87 / 35.94 & 42.82 / 24.99 &34.65 / 28.62 & 117.46 / 31.94 & 130.61 / 25.24 & 31.63 / 27.02 & 97.64 / 33.00
\\\bottomrule
\end{tabular}
}
\caption{Statistics of \dsname{}. For each language pair, we present the number of phrase pairs, average phrase length, and average length of example sentences. Notice that the length is counted by characters for Japanese (ja), Korean (ko), and Chinese (zh), and counted by words for the other languages.}
\label{table:ds-detail}
\end{table*}

\begin{table*}[htbp]
\centering
\scalebox{0.9}{
\begin{tabular}{lcccccccc}
\toprule
\textbf{src$\backslash$ trg} &\textbf{ar-en}& \textbf{de-en} & \textbf{en-es} & \textbf{en-fr} & \textbf{en-ja} & \textbf{en-ko}& \textbf{en-ru} & \textbf{en-zh}   \\ \midrule
\bf ar-en & 88.63 & 68.01 & 79.55 & 73.52 & 63.82 & 73.82 & 83.58 & 55.74  \\
\bf de-en & 53.80 & 81.44 & 84.16 & 78.84 & 46.93 & 61.92 & 70.52 & 47.61  \\
\bf en-es & 51.96 & 70.99 & 84.53 & 79.15 & 42.47 & 57.76 & 67.65 & 43.60 \\
\bf en-fr & 51.15 & 71.17 & 83.60 & 80.18 & 41.40 & 57.20 & 65.88 & 43.16   \\
\bf en-ja & 78.93 & 76.19 & 82.70 & 77.85 & 87.32 & 84.66 & 87.25 & 74.24 \\
\bf en-ko & 67.41 & 66.05 & 76.78 & 72.34 & 62.34 & 80.83 & 77.62 & 56.93 \\
\bf en-ru & 73.48 & 70.58 & 81.23 & 76.03 & 64.33 & 73.58 & 91.00 & 54.62 \\
\bf en-zh & 74.12 & 73.60 & 82.54 & 77.36 & 73.04 & 78.52 & 79.10 & 77.62  \\

\bottomrule
\end{tabular}
}
\caption{Evaluation results of \our{} on \dsname{} under the zero-shot transfer setting. `src' denotes the source language pair for training. `trg' denotes the target language pair for evaluation.}
\label{table:zsxpr}
\end{table*}

\end{document}